\documentclass[12pt]{article}

\usepackage{newtxtext,newtxmath}
\usepackage{graphicx}
\usepackage{xspace}
\usepackage[letterpaper,margin=1in]{geometry}

\renewenvironment{abstract}
	{\quotation}
	{\endquotation}

\date{}

\makeatletter
\renewcommand{\fnum@figure}{\textbf{Figure \thefigure}}
\renewcommand{\fnum@table}{\textbf{Table \thetable}}
\makeatother

\usepackage{scicite}

\usepackage{url}

\def\scititle{
	Transferring Expert Cognitive Models to Social Robots via Agentic Concept Bottleneck Models
}
\title{\bfseries \boldmath \scititle}

\author{
	Xinyu~Zhao$^{1\dagger}$,
        Zhen~Tan$^{2\dagger}$,
        Maya~Enisman$^{3}$,
        Minjae~Seo$^{3}$\and
        Marta~R.~Durantini$^{3}$,
        Dolores~Albarracin$^{3}$,
	Tianlong~Chen$^{1\ast}$\and
	\small$^{1}$The University of North Carolina at Chapel Hill\and
	\small$^{2}$Arizona State University\xspace
	\small$^{3}$University of Pennsylvania\\
	\small$^\ast$Corresponding author. Email: tianlong@cs.unc.edu\\
	\small$^\dagger$These authors contributed equally to this work.
}

\begin{document} 

% Insert the title and author list
\maketitle

% Abstract, in bold
% There are strict length limits, and not all formats have abstracts.
% Consult the journal instructions to authors for details.
% Do not cite any references in the abstract.
\begin{abstract} \bfseries \boldmath
% Start with one or two sentences of background
Successful group meetings, such as those implemented in group behavioral-change programs, work meetings, and other social contexts, must promote individual goal setting and execution while strengthening the social relationships within the group. Consequently, an ideal facilitator must be sensitive to the subtle dynamics of disengagement, difficulties with individual goal setting and execution, and interpersonal difficulties that signal a need for intervention. The challenges and cognitive load experienced by facilitators create a critical gap for an embodied technology that can interpret social exchanges while remaining aware of the needs of the individuals in the group and providing transparent recommendations that go beyond powerful but ``black box'' foundation models (FMs) that identify social cues. We address this important demand with a social robot co-facilitator that analyzes multimodal meeting data and provides discreet cues to the facilitator. The robot’s reasoning is powered by an agentic concept bottleneck model (CBM), which makes decisions based on human-interpretable concepts like participant engagement and sentiments, ensuring transparency and trustworthiness. Our core contribution is a transfer learning framework that distills the broad social understanding of an FM into our specialized and transparent CBM. This concept-driven system significantly outperforms direct zero-shot FMs in predicting the need for intervention and enables real-time human correction of its reasoning. Critically, we demonstrate robust knowledge transfer: the model generalizes across different groups and successfully transfers the expertise of senior human facilitators to improve the performance of novices. By transferring an expert’s cognitive model into an interpretable robotic partner, our work provides a powerful blueprint for augmenting human capabilities in complex social domains.
\end{abstract}

% The first paragraph of any Science paper does NOT have a heading
% Nor is it indented
\noindent
Effective human interaction that promotes both individual and social goals is inherently difficult for both groups and those facilitating a meeting. Also, a substantial portion of modern collaboration, therapy, and education now occurs through digital interfaces~\cite{belpaeme2018social,mataric2017socially,chen2025social}, adding to the challenges for effective human interaction that promote both individual and social goals~\cite{kragic2021effective}. While these digital platforms increase accessibility, they filter out the rich, non-verbal stream of social data—gaze, posture, and sentiment—that humans intuitively use to build trust and rapport~\cite{steiner1997essentials,ghonasgi2024crucial}. Along with the extreme challenges of leading goal-driven meetings, this perceptual gap places an immense cognitive load on facilitators, teachers, and therapists, who must manage group dynamics with incomplete information~\cite{cravens2022science,gearhart2022barriers}. To fulfill these complex social roles, intelligent systems must be fast and effective at interpreting verbal content while also being able to perceive and act upon subtle, implicit social cues without forgetting that meetings must achieve individual and collective goals.

Embodied robotic agents offer a comprehensive solution that addresses these requirements. Unlike disembodied AI assistants such as Otter.ai~\cite{corrente2022innovation}, which excel at transcribing dialogue but remain blind to its social context, a physically co-located robot can perceive the full multimodal data stream of an interaction. However, fully leveraging such machines in autonomous, high-stakes applications, such as behavioral interventions and psychotherapy, requires solving several fundamental challenges. First, transferring human expertise remains a difficult and open issue. The nuanced, context-sensitive intuition of a skilled facilitator is not easily codified into rules or policies~\cite{collins2019tacit,howells1996tacit,caby2023techniques}. Determining how to capture this expert knowledge and embed it within a robotic system is not straightforward, as successful intervention depends heavily on subtle cues that are difficult to label and model. Second, the emergence of large-scale foundation models (FMs) presents a new paradigm for understanding social data~\cite{grossmann2023ai,ashery2025emergent}, yet their ``black box'' nature poses a major obstacle to deployment~\cite{hassija2024interpreting}. For a robotic assistant to be trusted in sensitive human environments, its reasoning must be transparent, auditable, and predictable. End-to-end models that produce actions from raw sensor data cannot provide these guarantees, limiting their utility where safety and accountability are paramount. Third, attaining effective human-robot teaming poses a substantial engineering challenge, requiring seamless integration between perception and action~\cite{roesler2021meta}. A robot cannot be a passive observer; it must become an active participant in the social loop. Issues observed in disjointed systems include slow response times or actions that are not appropriately synchronized with the evolving social context, which can compromise the very dynamic the robot is intended to support.

In this work, we developed a socially assistive robotic system that enables seamless coordination between social perception and facilitator guidance. Our approach integrates a novel transfer learning framework that distills the broad social understanding of a general foundation model into an interpretable and agentic Concept Bottleneck Model (CBM)~\cite{koh2020concept}. This framework first deploys LLMs to map high-dimensional sensory data onto a set of intermediate, human-understandable concepts, such as participant ``engagement'', ``interaction'', and ``sentiment'', before making a final prediction. This two-stage process makes the robot’s reasoning transparent and allows a human to intervene by correcting concept-level predictions. We demonstrate our system in the challenging domain of an online group behavioral intervention~\cite{bright2023mental,bantjes2021web}, where a social robot acts as a co-facilitator. Extensive evaluation shows that our system successfully predicts the need for intervention, allows for real-time human correction of its reasoning, and robustly transfers knowledge between different groups and facilitator experience levels. Our practical evaluations emphasize the potential of this architecture for achieving transparent and effective human-robot collaboration in real-world applications.

% Research Articles and Reviews split the text into sections using headings
% Use a short (up 6 words) descriptive phrase, not generic 'Results' or 'Conclusions'
% Most other formats do not have headings, see the journal instructions to authors for details
\begin{figure}
	\centering
	\includegraphics[width=0.7\textwidth]{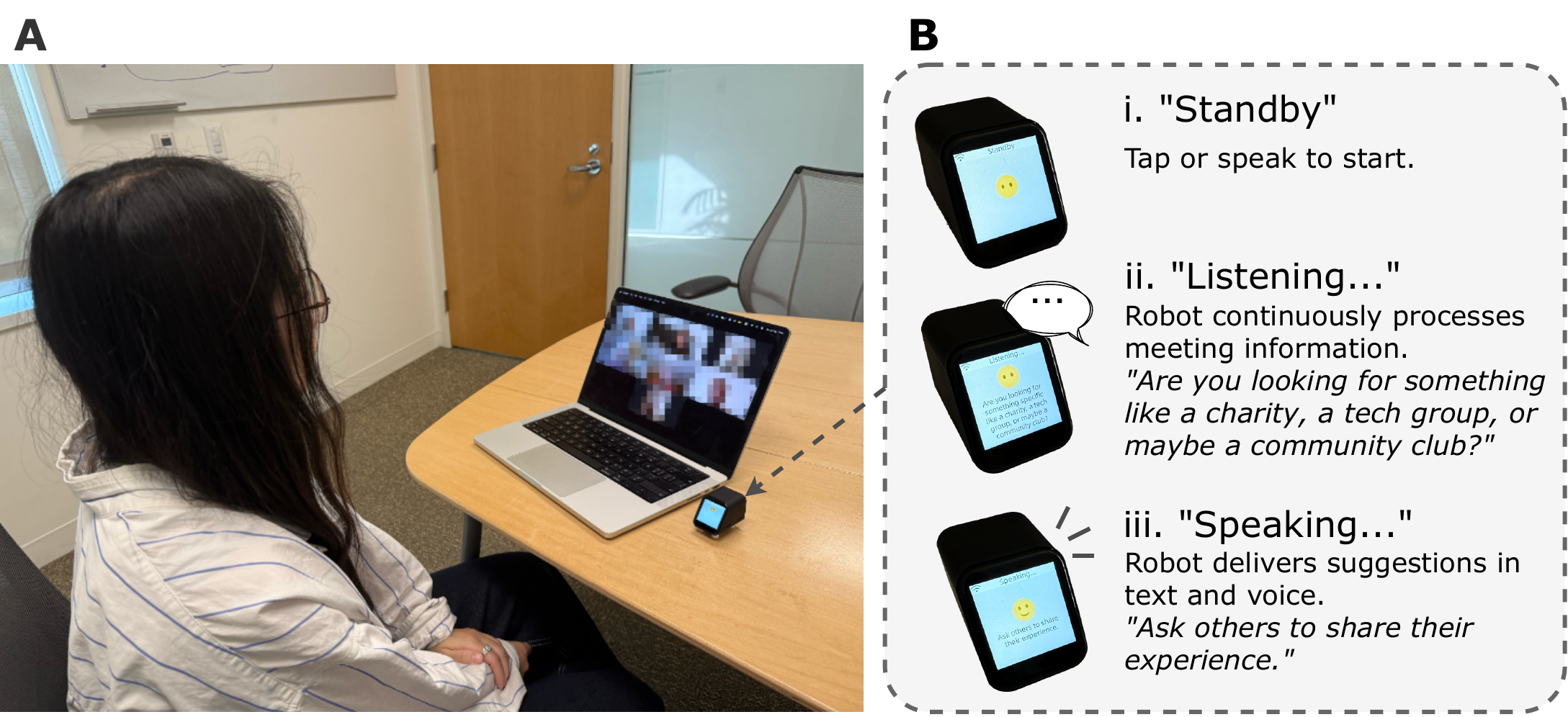} 
	\caption{\textbf{
		Deployment and interaction workflow of the social robot facilitator agent. } The system is designed to provide discreet, real-time assistance to a human facilitator during virtual group intervention sessions. \textbf{(A)} A representative use case of the system. A facilitator engages and monitors a group meeting on a laptop, with the social robot agent embodied as a peripheral device that delivers intervention suggestions. \textbf{(B)} The agent's three-stage operational loop: \underline{(i)} the agent is activated by the facilitator, \underline{(ii)} it continuously analyzes the meeting dynamics, and \underline{(iii)} it provides timely intervention suggestions directly to the facilitator.}
	\label{fig:example}
\end{figure}

\section*{Results}
\subsection*{Design for Group Intervention Meeting}
We conduct experiments to evaluate the efficacy of our multimodal chatbot, designed to assist facilitators by enhancing interaction dynamics in online group meetings akin to the settings in~\cite{bright2023mental,bantjes2021web}. The evaluation is performed on a dataset derived from 8 videos of online group interventions, from which 517 discrete segments are extracted and annotated by human experts.

The program is delivered to groups remotely using Zoom to ensure program delivery, wide accessibility, and sustainability. The intervention is founded on three premises. First, it seeks to increase social capital and reduce substance use prejudice by integrating people who use substances with the community at large~\cite{amaro2021social,davis2023community,lundborg2005social}. Second, it uses a participant-centered approach where the intervention adapts to the individual needs of attendees, promoting individual goal-setting achievement~\cite{lundborg2005social,bahrami2022applying}. Third, the program addresses health access by making resources like HIV and HCV testing kits and Narcan available remotely. Sessions are led by facilitators trained to manage a behavioral intervention that capitalizes on a supportive, generative group dynamic and psychotherapy techniques like goal setting, homework, reframing, and incisive question asking. They are also prepared with resolutions for potential adverse events and equipped to handle situations where a participant expresses distress to safeguard all participants. The intervention includes three 90-minute sessions taking place over three weeks. The first session is designed to foster individual goals to reconnect with lost family members or friends, or to create new social connections. The second session is designed to promote each individual’s physical or mental health goals. The third session is designed to promote each individual’s goals for the next 3 to 6 months.

\begin{figure}
	\centering
	\includegraphics[width=0.8\textwidth]{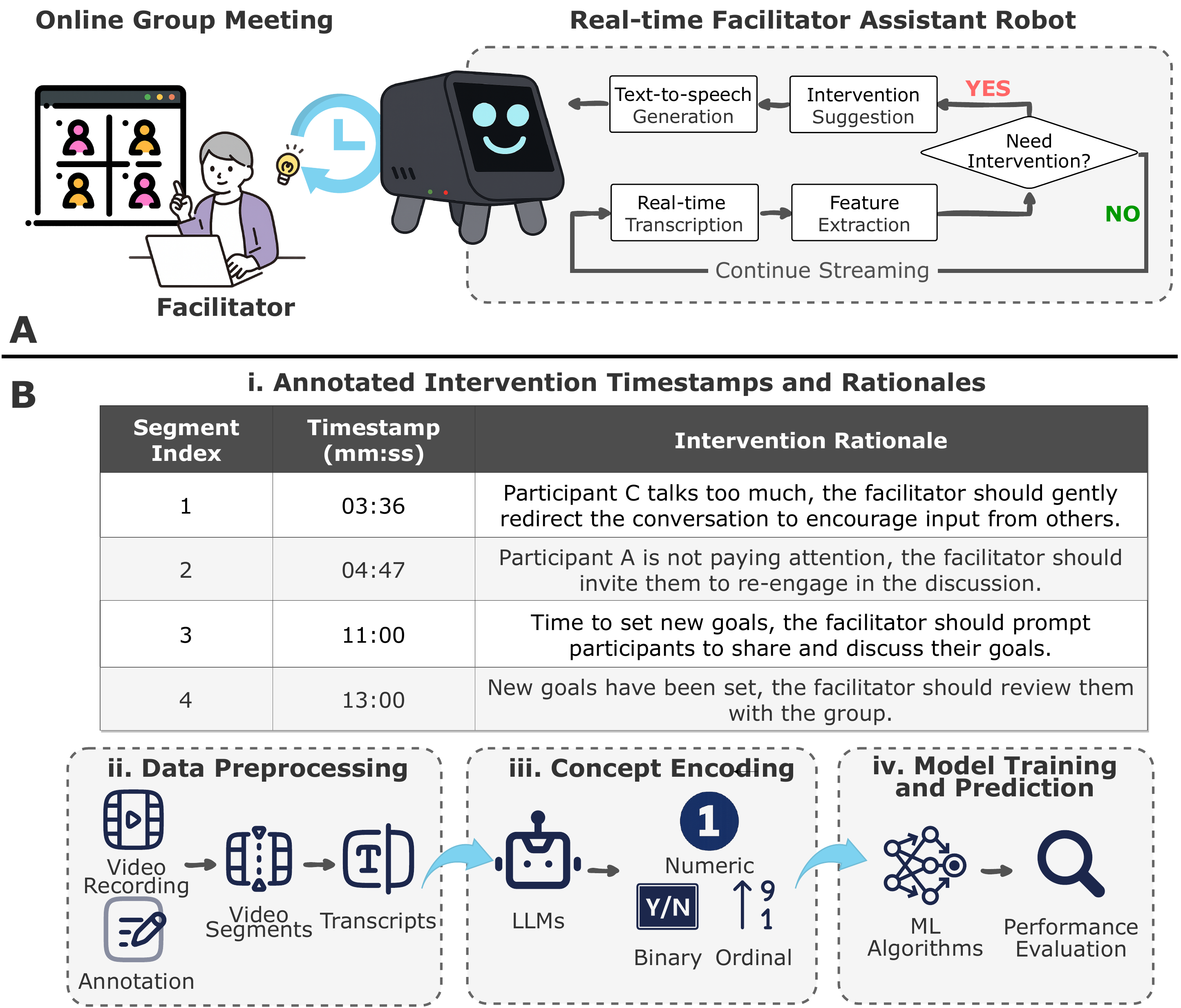}
	\caption{\textbf{
		Overview of the real-time facilitator assistant system for group intervention support.} \textbf{(A)} The human-robot interaction model. A social robot assists a facilitator during an online meeting by analyzing the interaction in real-time. If it predicts an intervention is needed, it delivers a discreet voice suggestion to the human operator. \textbf{(B)} The system’s machine learning pipeline. \underline{(i)}  Ground-truth data is established from expert annotations of intervention timestamps and rationales. \underline{(ii)} Raw video recordings are preprocessed into transcribed segments. \underline{(iii)} An LLM performs concept encoding, translating transcripts into a structured vector of interpretable concepts. \underline{(iv)} This concept vector is used to train and evaluate the final model that predicts the need for intervention.}
	\label{fig:deploy}
\end{figure}

To establish the ground truth for our predictive models, social science experts annotated these video recordings, identifying precise timestamps where an intervention was needed and providing a rationale. As shown in Fig.~\ref{fig:deploy}B, common intervention scenarios included moments where a participant was talking too much, when a participant was not paying attention, or when it was time for individuals to set up and make plans to achieve an important and feasible goal. Following these expert annotations, we extracted 517 discrete 60-second segments, resulting in 358 positive samples (requiring intervention) and 159 negative samples (not requiring intervention).

The overall system pipeline involves pre-processing the video recordings into segments and transcripts, using LLMs for concept encoding, and then training and evaluating a machine learning model for prediction. Based on this structure, we formulate our core predictive challenge as a dual task. For each meeting segment, the system must first predict whether an intervention is needed. This binary classification task determines if the situation warrants facilitator action. If an intervention is warranted, the system must then generate a concrete and actionable recommendation, which the robot can deliver to the human facilitator as either a voice or text suggestion, based on the facilitator's preference.

\subsection*{Research Questions}
Our experimental design is structured to address three research questions. 

The first research question (\textbf{RQ1}) examines the effectiveness of our agentic concept bottleneck model (CBM) in predicting the need for facilitator intervention. Foundation models (FMs) show promise in understanding social data, but their ``black box'' nature limits their use in high-stakes, sensitive environments where trust and transparency are critical. This question addresses the core challenge of whether our proposed framework—which distills knowledge from a general FM into a specialized, transparent CBM—can achieve superior performance compared to directly using the FM in a zero-shot or few-shot capacity. Through this comparison, we seek to validate the benefits of our structured, concept-driven architecture for reliable decision-making.

The second research question (\textbf{RQ2}) investigates the system’s ability to generate actionable and contextually relevant intervention recommendations. A successful co-facilitator must do more than simply identify when to intervene; it must provide guidance that is specific, timely, and aligned with expert facilitation practices. This question evaluates whether our system can synthesize its understanding of session history, real-time dynamics, and procedural goals to produce high-quality, human-like advice that can effectively support a facilitator.

The third research question (\textbf{RQ3}) explores how the system’s transparent architecture enables real-time, human-in-the-loop enhancement. For a robotic assistant to be a true partner, its reasoning must be auditable and correctable. This question examines the practical utility of the CBM’s interpretable concepts, assessing whether a human facilitator can intervene at the concept level to correct the model’s reasoning and improve its predictive accuracy during test time. This question investigates the potential for a collaborative partnership between the human and the robot, a critical factor for deployment in complex social domains.

Grounded in the limitations of direct FM application and the theoretical benefits of interpretable models, we hypothesized that our proposed framework would yield more positive results than baseline approaches. Specifically, we hypothesized that the CBM-based predictor would significantly outperform direct zero-shot and few-shot FM predictions (\textbf{RQ1}). We further hypothesized that the system’s multi-faceted understanding would enable it to generate high-quality, actionable suggestions (\textbf{RQ2}), and that its transparent conceptual layer would allow for effective human-in-the-loop corrections to improve accuracy (\textbf{RQ3}).

\begin{figure}
	\centering
	\includegraphics[width=0.8\textwidth]{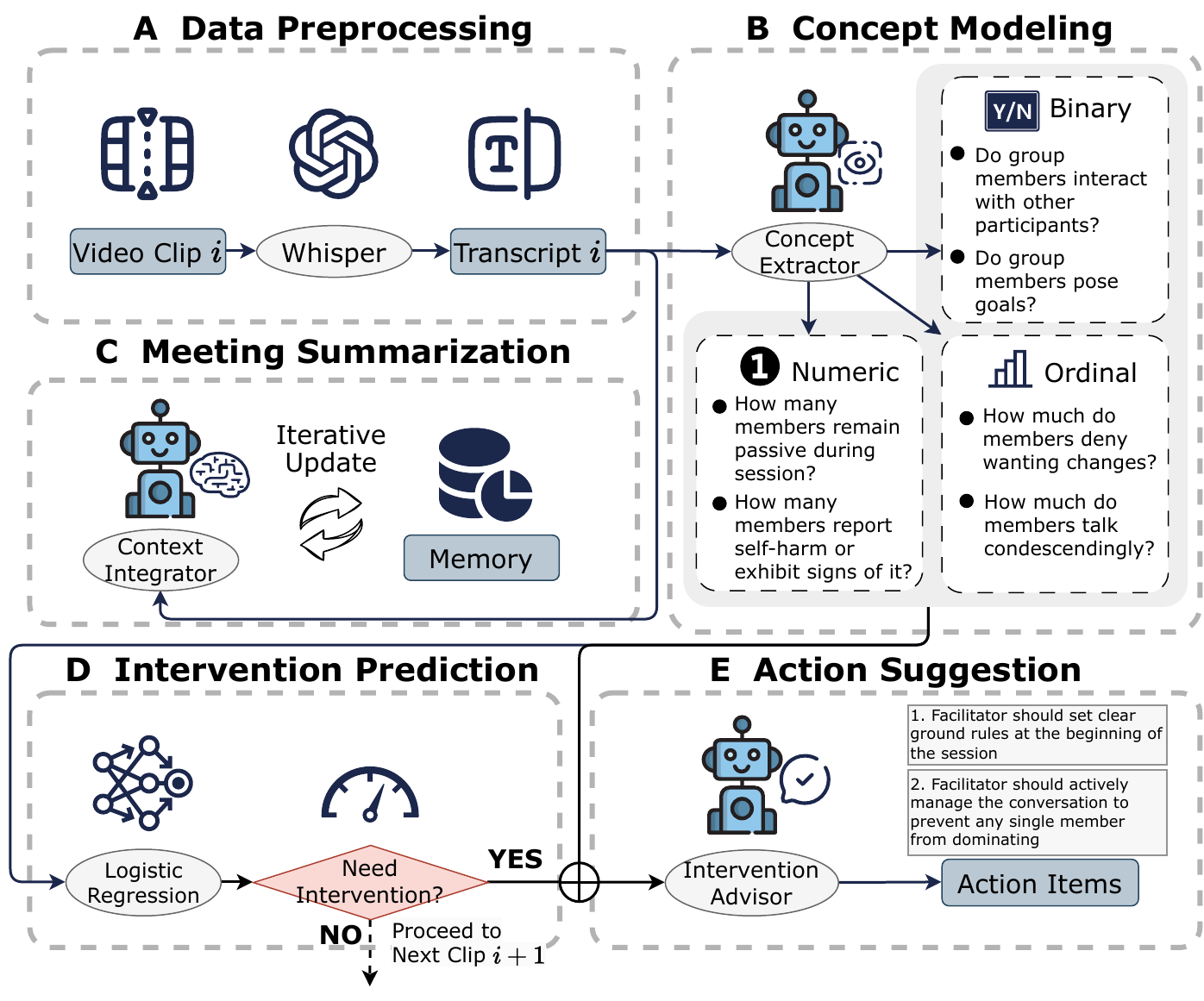}  
	\caption{\textbf{
		Deployment and interaction workflow of the social robot facilitator agent.  The system is designed to provide discreet, real-time assistance to a human facilitator during virtual group intervention sessions.} The robot demonstrates a multifaceted understanding of the meeting by providing nuanced advice tailored to different situations. \textbf{(A)} A representative use case of the system. A facilitator engages and monitors a group meeting on a laptop, with the social robot agent embodied as a peripheral device that delivers intervention suggestions. \textbf{(B)} The agent's three-stage operational loop: \underline{(i)} the agent is activated by the facilitator, \underline{(ii)} it continuously analyzes the meeting dynamics, and \underline{(iii)} it provides timely intervention suggestions directly to the facilitator.}
	\label{fig:method}
\end{figure}

\subsection*{System Overview}
We present a multi-agent system designed to understand group dynamics and assist human facilitators. The system architecture, illustrated in Fig.~\ref{fig:method}, comprises several interconnected modules that handle data pre-processing, concept modeling, context summarization, and finally intervention prediction and suggestion.

\subsubsection*{Data Annotation and Pre-processing}
The foundation of our system is a custom dataset curated from video recordings of real-world online group intervention sessions. To prepare this data for model training and evaluation, we implemented a structured pre-processing pipeline involving transcription, expert annotation, and systematic sample generation.

The source data consists of full-length video recordings of group meetings. First, each video was transcribed using the OpenAI Whisper model to generate time-stamped conversational text. To establish a ground truth for our predictive task, a social science expert then performed a detailed annotation of each session. Following established qualitative coding practices, the expert created a coding sheet identifying the precise timestamps where a facilitator intervention was deemed necessary and provided a detailed rationale for each decision.

From the annotated recordings, we generated discrete data samples for training and validation. This procedure was designed to create a balanced set of examples representing moments that either required intervention (positive samples) or did not (negative samples). To create positive samples whose segments, require intervention, we extracted video and transcript clips corresponding to expert-annotated timestamps. Based on the principle that the need for an intervention becomes most salient in the period immediately preceding the event, we defined each positive sample as the 60-second window leading up to and including the annotated time of need. To create negative samples whose segments do not require intervention, we identified long, continuous periods in the session recordings where no intervention codes were present. Specifically, any continuous segment between two intervention codes with a gap of five minutes or more was considered a source of negative data. These uneventful periods were then partitioned into multiple non-overlapping 60-second chunks to serve as negative samples.

\subsubsection*{Assisting Intervention Decision-making and Suggestion}
After a comprehensive understanding of the meeting state has been concluded, the system transitions to its assistive role. The assisting process involves two sequential modules: Intervention Prediction, which determines if a facilitator’s intervention is needed, and Action Suggestion, which generates specific advice if an intervention is needed.

To provide timely support, the robot must first predict whether the current state of the group interaction requires facilitator intervention. We formulate this as a binary classification task. Following the structure of a concept bottleneck model, this module uses the structured concepts predicted by the Concept Extractor as its direct input. Let the vector of k predicted concepts for a given time segment be $\hat{c}\in\mathbb{R}^k$. The intervention prediction module learns a function $f(\hat{c})\ \rightarrow\hat{y}$, where $\hat{y}$ is the predicted probability that an intervention is needed. This function is implemented as a lightweight and computationally efficient classifier. The raw conceptual outputs from the Concept Extractor are first post-processed into a numerical feature vector suitable for the classifier. A Logistic Regression model is used for this task after a comparative cross-validation analysis. The output of this module is a binary decision, which determines whether the subsequent Action Suggestion module is triggered. If the decision is negative, the system proceeds to the next video clip. 

When the Intervention Prediction module identifies a need for intervention, the Intervention Advisor agent is activated. This agent is designed to provide the human facilitator with concrete, context-aware, and actionable suggestions. To generate high-quality advice, the agent leverages the full scope of the system’s understanding. Its decision is conditioned on a rich set of inputs: the holistic, cumulative summary of the meeting dynamics generated by the Context Integrator; the current session’s predefined stage-specific goals; and the raw transcript from the immediate time segment. This multifaceted context ensures that suggestions are not generic but instead tailored to the specific situation, its history, and its intended trajectory. The Intervention Advisor is also guided by a few-shot examples of expert-annotated interventions, each pairing a real-world transcript segment with an ideal recommended action and its rationale. This technique steers the model’s reasoning process, enabling it to generate suggestions that align with expert facilitation strategies. The final output is a structured object containing a short, recommended action and a detailed rationale, which explains why the action is necessary at that moment.

\begin{figure}
	\centering
	\includegraphics[width=\textwidth]{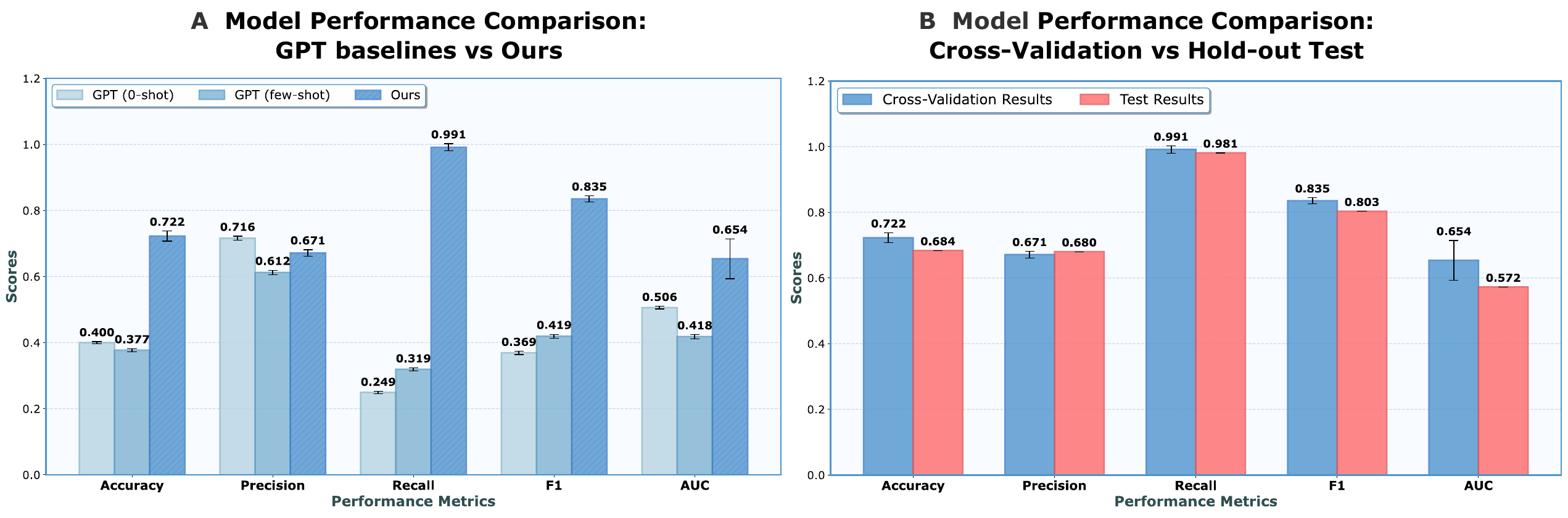}
	\caption{\textbf{Intervention prediction performance.} \textbf{(A)} Comparison of our proposed model against GPT (0-shot) and GPT (few-shot) baselines for predicting the need for facilitator intervention. Our model, which uses a CBM architecture, demonstrates substantially higher performance across all metrics, most notably achieving near-perfect recall. \textbf{(B)} Comparison of our model's performance under 5-fold cross-validation versus a held-out test set, demonstrating robust generalization with minimal performance degradation. }
	\label{fig:quantitative}
\end{figure}

\subsection*{RQ1: How Does the Social Robot Assist Intervention Decision-Making?}
First, we analyze the ability of our multi-agent system to predict the need for facilitator interventions during group meetings. We implemented the multi-agent system with all agents using GPT-4 as the backbone model. The resulting conceptual feature vectors were then used to train and evaluate a logistic regression classifier with an elastic net penalty and an inverse of regularization strength of 1.0, which was weighted to account for potential data imbalance. We evaluated the classifier using a 5-fold cross-validation scheme to ensure a robust estimation of its effectiveness. For comparison, we established two baseline conditions using the same backbone GPT-4: a zero-shot model that received only the raw transcript, and a few-shot model that was also provided with expert-annotated examples. The performance of the baselines was averaged over five independent runs, and all conditions were measured using Accuracy, Precision, Recall, F1-score, and Area Under the Curve (AUC) for the Receiver Operating Characteristic curve (ROC).

The results, shown in Figure~\ref{fig:quantitative}A, reveal that the proposed agentic concepts bottleneck framework substantially outperforms direct prompting of the LLM. Our model, which first translates conversational data into high-level concepts before classification, achieves high performance across all metrics. Most notably, the model achieved a near-perfect recall score (0.991), showcasing its ability to successfully identify nearly every instance where a facilitator’s intervention is required. In a support application such as this, high recall is critical, as it ensures the system does not fail to alert the facilitator when assistance is needed. In contrast, both the zero-shot and few-shot GPT-4 baselines performed poorly, with the addition of few-shot examples yielding no significant improvement. This limitation demonstrates that even a powerful foundation model struggles with the nuance of this task when prompted directly, highlighting the necessity of our structured, concept-driven approach.

To further evaluate the model’s robustness, we tested it on a held-out test set. As illustrated in Figure~\ref{fig:quantitative}B, the model maintains excellent performance, demonstrating robust generalization with minimal degradation on unseen data. This finding indicates that the conceptual knowledge learned by our model is stable and can be reliably transferred to new sessions.

\begin{figure}
	\centering
	\includegraphics[width=\textwidth]{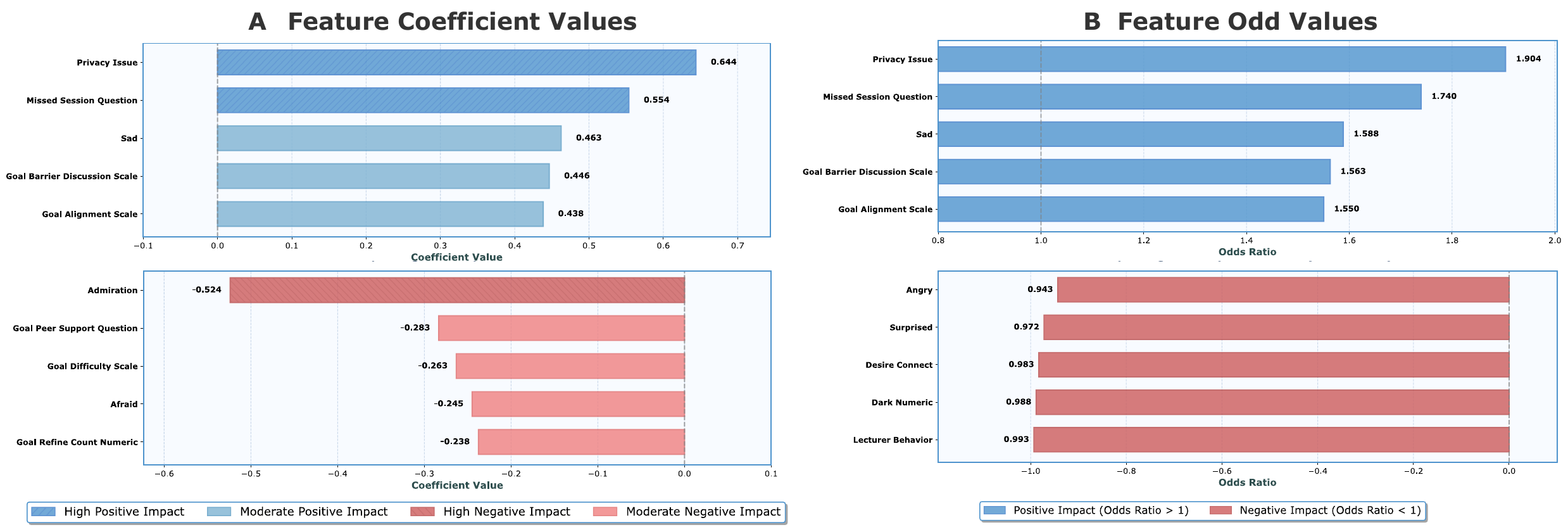}
	\caption{\textbf{Intervention feature analysis.} \textbf{(A)} Feature coefficient values from the trained logistic regression model, indicating the weight and direction of the influence of the concepts on the prediction. Positive values (blue) are associated with a higher likelihood of intervention. \textbf{(B)} Feature odds ratios, illustrating the multiplicative effect on the odds of intervention for a one-unit increase in each concept. Values over 1.0 indicate a positive association with the need for intervention.}
	\label{fig:feature_analysis}
\end{figure}

\subsubsection*{Feature Importance Analysis}

To understand why the model makes its decisions, we analyzed the feature coefficients and odds ratios from the trained logistic regression model (Fig.~\ref{fig:feature_analysis}). This analysis provides a transparent view into the model’s reasoning, revealing which concepts are most predictive of the need for an intervention and demonstrating a strong alignment with expert facilitation practices.

The concepts with positive coefficients increase the likelihood of the model recommending an intervention. The most significant predictor is ``Privacy Issue'', referring to the presence of another adult in the participant’s space, with the largest positive coefficient (0.644) and odds ratio (1.984). This indicates that when privacy concerns arise, the odds of suggesting an intervention nearly double. Other strong positive predictors include: ``Missed Session Question'' (coefficient: 0.554; odds ratio: 1.741), capturing attempts to re-engage with group norms; ``Sad'' (0.463; 1.589), indicating emotional vulnerability; and ``Goal Barrier Discussion Scale'' (0.446; 1.562), reflecting deeper engagement with goal-related challenges. These results suggest the model prioritizes moments marked by emotional distress, disrupted participation, or critical reflection on behavioral goals.

Conversely, concepts with negative coefficients decrease the likelihood of an intervention. These include indicators of active engagement and positive tone, such as ``Admiration'' ($-0.524$), ``Goal Peer Support Question'' ($-0.283$), and ``Goal Refine 
Count'' ($-0.238$). This suggests the model avoids interrupting productive group dynamics. Some negative predictors appear counterintuitive, such as ``Goal Difficulty Scale'' and the emotion ``Afraid.'' This suggests that while these signals may reflect individual struggle, they are not, in isolation, strong enough to warrant an intervention recommendation.Besides, there are other negative indicators not aligned with intuition, such as ``Goal Difficulty Scale'' and ``Afraid'' emotion. This result suggests that while these behaviors are monitored, their isolated presence is not sufficient to trigger an intervention recommendation. All these detailed feature analysis confirms that the decision-making for intervention is not only accurate but also grounded in a logical, interpretable, ethical, and contextually relevant understanding of group dynamics that mirrors human expertise.

\begin{figure}
	\centering
	\includegraphics[width=\textwidth]{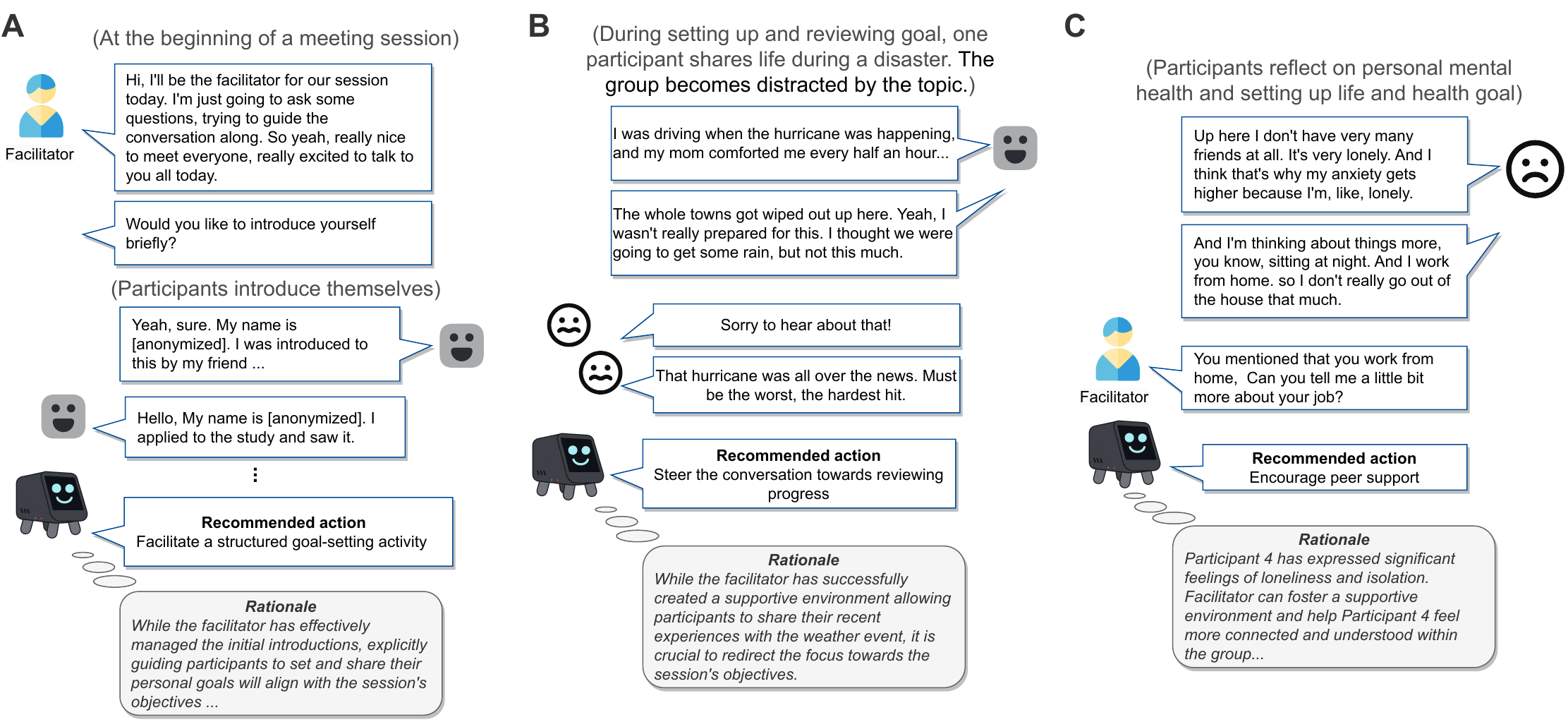}
	\caption{\textbf{
		Examples of context-aware intervention suggestions generated by the social robot.} The robot demonstrates a multifaceted understanding of the meeting by providing nuanced advice tailored to different situations. \textbf{(A)} The robot proposes a ``goal intervention'', prompting the facilitator to begin a structured activity after introductions. \textbf{(B)} The robot suggests a ``redirect intervention'', advising the facilitator to steer a distracting conversation back to the agenda. \textbf{(C)} The robot recommends a ``support intervention'' by prompting the encouragement of peer support after a participant expresses vulnerability.}
	\label{fig:qualitative}
\end{figure}

\subsection*{RQ2: How to Enable the Social Robot to Generate Actionable Intervention Recommendations}
Next, we qualitatively evaluate the social robot’s ability to generate actionable, contextually relevant, and procedurally appropriate intervention suggestions. We assess whether the Intervention Advisor agent can synthesize its multifaceted understanding of the meeting to produce high-quality, human-like advice. The recommendations are generated by the Intervention Advisor across a range of representative meeting scenarios. The agent is prompted with the cumulative session summary, concepts from the current time segment, the session’s predefined procedural goals. and the raw transcript. Its outputs, consisting of a recommended action and a rationale, are evaluated for their relevance, specificity, and alignment with expert facilitation practices.

Our findings indicate that the agent architecture enables the social robot to generate nuanced and highly relevant intervention recommendations that align with expert strategies. By integrating its different modes of understanding, the agent demonstrates three key capabilities, as illustrated in Fig.~\ref{fig:qualitative}. First, the agent is procedure-aware. At the beginning of a meeting, after introductions are complete, the agent identifies the need to move to the next item on the agenda. It suggests a ``goal-setting'' intervention to ``Facilitate a structured goal-setting activity'' (Fig.~\ref{fig:qualitative}A). This recommendation directly reflects its knowledge of the predefined session structure, ensuring the meeting moves toward its objectives. 

Second, the agent effectively grasps in-the-moment dynamics. When a participant’s personal story about a disaster distracts the group, the agent detects the deviation from the topic and recommends a 'redirect' intervention: to ``Steer the conversation towards reviewing progress'' (Fig.~\ref{fig:qualitative}B). This instance shows the agent’s ability to use contextual cues to suggest immediate actions that restore the group’s focus on the task. 

Third, the agent is context-aware and can interpret emotional subtexts. After a participant shares a vulnerable experience of loneliness, the agent recognizes the emotional weight of the moment and suggests a 'support' intervention, advising the facilitator to ``Encourage peer support'' (Fig.~\ref{fig:qualitative}C). This advice demonstrates a deeper level of comprehension, moving beyond literal content to suggest interventions that foster group cohesion and empathy.

\begin{figure}
	\centering
	\includegraphics[width=0.7\textwidth]{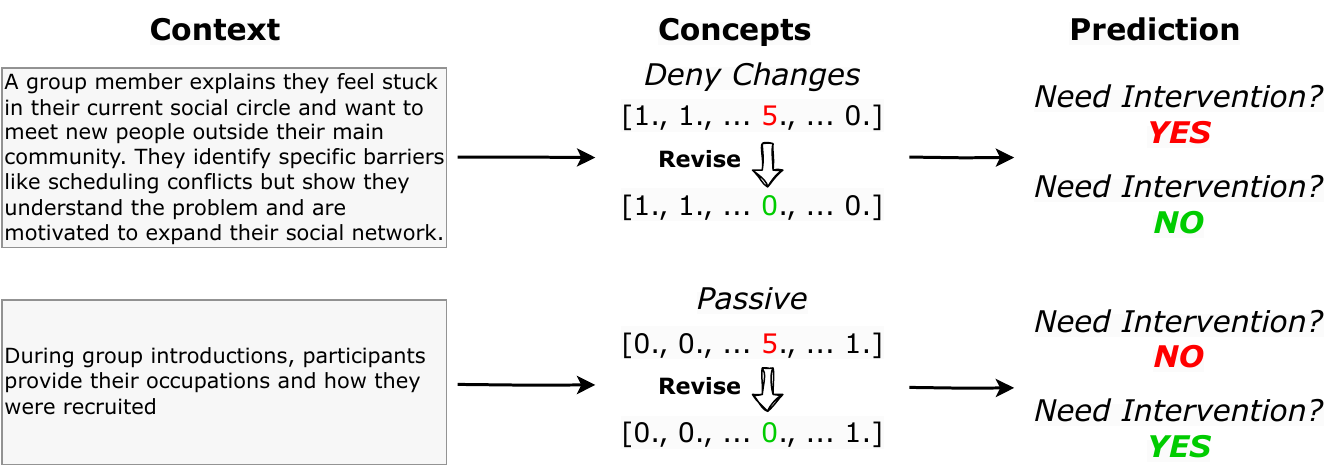}
	\caption{\textbf{
		Example of test-time concept editing of the social robot to correct model predictions.} This figure illustrates the human-in-the-loop capability of the CBM architecture. An incorrect concept prediction by the model can be manually revised by the facilitator, leading to a corrected final prediction. (Top) The model initially misinterprets a participant’s motivated problem-solving as “Denying Change”, leading to an incorrect "YES" prediction for intervention. A human corrects the Deny Changes concept from 5 to 0, which flips the final prediction to the correct “NO”. (Bottom) The model mislabels active introductions as “Passive”, leading to an incorrect “NO” prediction. When a human corrects the Passive concept from 5 to 0, the model’s final prediction correctly changes to “YES”, signaling that the introductions are a key procedural step requiring facilitation.}
	\label{fig:tti}
\end{figure}

\subsection*{RQ3: How to Enhance the Social Robot via Test-Time Concept Editing}

A key benefit of the concept bottleneck architecture is its inherent interpretability, which enables human-in-the-loop corrections at test time. This analysis investigates whether manually correcting erroneously encoded concepts can rectify the model's final prediction, thereby improving its accuracy and demonstrating a pathway for effective human-robot collaboration.

We selected cases from our test set where the classifier initially made an incorrect prediction. For each case, we identified the specific concept that the concept extractor agent had miscategorized. We then simulated a human-in-the-loop intervention by manually revising the value of the incorrect concept in the feature vector to its ground-truth value. This corrected feature vector was then passed to the trained classifier to generate a new prediction, which was compared against the original output.

Our findings demonstrate that test-time editing of concepts is a powerful mechanism for correcting the model’s final prediction, highlighting its interactive and interpretable nature. As shown in the illustrative examples in Fig.~\ref{fig:qualitative}, this process can rectify distinct types of prediction failures.

In the first scenario (Fig.~\ref{fig:tti}, Top), a participant explains that they feel stuck but also identifies specific barriers and shows motivation to change. The Concept Extractor misinterprets the discussion of barriers as resistance, incorrectly assigning a high score (5) to the Deny Changes concept. This mislabeling leads to an erroneous prediction that an intervention is needed. When a human operator revises this single concept value to 0 to reflect the participant’s true motivation, the model’s prediction correctly flips to ``NO'', avoiding unnecessary interruption.

In the second scenario (Fig.~\ref{fig:tti}, Bottom), during group introductions, participants are mislabeled as being highly ``Passive'' (score of 5) for providing brief, standard answers. This mislabeling leads to an incorrect prediction that no intervention is needed. However, expert facilitation guidelines suggest that initial introductions are a key moment to encourage deeper engagement. The participants are not being truly passive but are awaiting facilitator guidance. Correcting the Passive concept to 0 allows the model to recognize this procedural nuance, rightly changing the prediction to ``YES'', and prompting the facilitator to engage the group more deeply. These examples confirm that the CBM architecture not only provides transparency but also creates a direct, effective channel for human-robot collaboration, allowing a human’s contextual expertise to instantly refine and improve the robotic assistant’s performance.

\section*{Discussion} 
Our work demonstrates that by transferring knowledge from a general foundation model into an interpretable, agentic concept bottleneck model, a social robot can effectively assist human facilitators in the complex, high-stakes domain of online group intervention. The proposed framework significantly outperforms direct zero-shot or few-shot applications of foundation models in predicting the need for intervention (\textbf{RQ1}), generates nuanced, context-aware suggestions aligned with expert strategies (\textbf{RQ2}), and provides a transparent architecture amenable to human-in-the-loop correction (\textbf{RQ3}). The implications of this approach extend beyond the immediate application, offering a viable blueprint for human-robot teaming in sensitive social contexts and providing insights into the fundamental questions facing both social science and robotics.

\subsection*{Empowering Group Leaders’ Education and Training}
A significant implication of our findings is the potential to empower the education and training of new facilitators and meeting leaders. The intuition of an expert facilitator; knowing precisely when and how to intervene to manage group dynamics, de-escalate conflict, or foster vulnerability; is a skill developed over years of practice and is notoriously difficult to codify and teach. Our system addresses this challenge directly by capturing the cognitive model of an expert and making it accessible to a novice in real-time. As demonstrated in our results (\textbf{RQ1}, \textbf{RQ2}), the robot, armed with a senior facilitator’s distilled knowledge, can provide prompts and rationales that a novice might not have considered. This ability transforms the assistive robot into a pedagogical tool, providing a ``cognitive scaffold'' that guides trainees through live sessions. It effectively allows novices to ``borrow'' an expert’s intuition, accelerating their learning curve, improving the quality of care they provide from the outset, and building their confidence in managing challenging interactions.

\subsection*{Supporting Equitable and Accessible Social Support}
The challenge of providing equitable access to mental health and social support resources is a pressing global issue, often limited by the availability of highly skilled professionals. Our transfer learning framework offers a scalable solution to this problem. The expert cognitive model, once distilled into the CBM, is not confined to a single location or person. It can be deployed across numerous robotic systems, enabling skilled assistance in geographically remote, underserved, or under-resourced communities where access to trained therapists or social workers is scarce. By embedding expertise into accessible technology, this work paves the way for a new model of social support delivery. It can augment the capabilities of local health workers, community leaders, and peer support specialists, democratizing access to high-quality intervention facilitation and thereby making social support more equitable and broadly available. This approach helps to overcome the inherent scalability limitations of traditional, one-on-one training models.

\subsection*{The Critical Role of Interpretability for Human-Robot Teaming}
For a social robot to be accepted in high-stakes human environments like therapy or education, trust is paramount. Our findings for \textbf{RQ3} underscore that the CBM architecture is fundamental to building trust. Unlike opaque, end-to-end models whose decision-making processes are inscrutable, our system’s reasoning is transparent by design. By exposing its intermediate concepts; participant passivity, conversational monopolization, goal setting; the robot allows the human facilitator to understand why a suggestion is being made. This interpretability is not a post-hoc feature but a core component of the robot's reasoning. It provides the crucial channel for the human-in-the-loop correction demonstrated in \textbf{RQ3}, allowing a facilitator to audit the robot’s ``thought process'' and correct it if a concept is misidentified. This collaborative dynamic, where the human’s contextual expertise can instantly refine the model’s judgment, is essential for fostering the trust, safety, and accountability required for effective human-robot teaming in complex social domains.

\subsection*{A New Paradigm for Computational Social Science}
Beyond its assistive capabilities, our methodology offers a new tool for social science research. Traditionally, studying complex group dynamics requires laborious, manual coding of interactions by trained researchers; a process that is slow, expensive, and difficult to scale. Our framework presents a paradigm for automating this crucial data analysis. By using an LLM to translate unstructured conversational data into a high-dimensional vector of expert-defined concepts, we effectively create a ``sociological lens'' through which vast amounts of interaction data can be analyzed quantitatively. This enables researchers to study the temporal evolution of group dynamics, identify subtle behavioral patterns that correlate with outcomes, and test social theories at an unprecedented scale. This work thus provides a blueprint for a new form of computational social science, where AI systems can serve not just as subjects of study, but as indispensable instruments for observation and discovery.

\subsection*{Bridging Symbolic and Sub-symbolic AI in Robotics}
From a robotics study perspective, our work contributes to a central challenge in the field: bridging the gap between sub-symbolic and symbolic AI. Sub-symbolic systems, like large neural networks, excel at perception and pattern recognition from raw data but lack transparency. Symbolic systems, based on logic and rules, are interpretable but often brittle and struggle with the ambiguity of the real world. Our CBM framework creates a robust bridge between these two paradigms. The LLM-based Concept Extractor acts as a powerful sub-symbolic engine, performing the perceptual heavy lifting of making sense of messy human language. The resulting concept vector is a structured, symbolic representation that is then used by a simple, auditable classifier. This hybrid architecture marries the perceptual power of deep learning with the clarity and safety of symbolic reasoning, offering a compelling model for building intelligent robots that are both highly capable and trustworthy.

\subsection*{What Should Be Transferred? From Policies to Conceptual Models}
This research provides a clear perspective on the foundational question of what can and should be transferred in robotics. While much of transfer learning has focused on transferring policies or low-level skills, our work argues that for complex social tasks, the most asset to transfer is an expert’s conceptual model of the world. Instead of learning a direct mapping from raw sensory data to an action, our system learns to first see the world through the eyes of an expert, translating messy, high-dimensional social data into the foundational concepts an expert would use to make a decision. The subsequent prediction task becomes vastly simpler and more robust. This concept-centric approach to transfer learning proved dramatically more effective than relying on the emergent, zero-shot capabilities of the foundation model alone. It suggests a powerful and generalizable paradigm for robotics, standing in contrast to end-to-end imitation learning methods that may fail unpredictably without offering recourse or explanation.

\subsection*{Limitations and Future Directions}
Despite the promising results, we acknowledge several limitations that provide avenues for future research. First, the set of concepts was defined by human experts; this is both a strength for interpretability and a potential bottleneck. The quality of the system is contingent on the quality of the expert-defined concepts. Future work could explore semi-supervised methods for discovering salient concepts directly from data, potentially revealing novel insights into group dynamics that even experts may overlook. Second, our current system relies solely on transcribed text. The rich social data present in visual cues (gaze, posture, facial expressions) and vocal prosody (tone, pitch, volume) are not yet integrated. Extending the CBM to be multimodal, with concepts derived from video and audio streams, is a clear and exciting next step that could further enhance the system’s perceptual capabilities and lead to more nuanced understanding. Finally, the system was evaluated in the context of online group meetings in a specific cultural setting. Future studies should investigate the adaptability and transferability of this framework across different cultural contexts, languages, and interaction scenarios, both online and in-person, to assess its broader applicability. Ethical considerations, including data privacy and the potential for misuse, must also be carefully addressed as these systems become more capable and widespread.

\section*{Materials and Methods}

\subsection*{Experiment Objective and Design}
The primary objective of this study was to design and evaluate a socially assistive robotic system that uses a novel transfer learning framework to act as a co-facilitator in online group interventions. The core of our approach is an agentic concept bottleneck model (CBM) that distills the broad social understanding of a general foundation model (FM), implemented by GPT-4 into a specialized, transparent, and interpretable model for predicting intervention needs. The experimental design was structured to answer three research questions: (\textbf{RQ1}) Determine the effectiveness of our agentic CBM in predicting the need for intervention compared to direct zero-shot and few-shot FM baselines. (\textbf{RQ2}) Qualitatively evaluate the system’s ability to generate actionable and contextually relevant intervention recommendations. (\textbf{RQ3}) Assess the potential for the system’s transparent architecture to be enhanced via real-time, human-in-the-loop concept correction.

\subsubsection*{Problem Formulation}
We formulate the assistive task as a dual challenge for each discrete time segment of a meeting. First, the system must perform a binary classification to predict whether a facilitator intervention is needed $y \in \left\{0,1\right\}$. This decision is learned by a function $f(\hat{c})\rightarrow\hat{y}$, where $\hat{c}\in\mathbb{R}^k$ is the vector of intermediate concepts extracted from the segment’s transcript. Second, if an intervention is predicted $\hat{y}\ =\ 1$, the system must perform a conditional generation task, producing a structured output containing a concrete, actionable suggestion for the human facilitator.	

\subsubsection*{Data Collection}
The dataset was derived from 8 full-length video recordings of real-world online group intervention meetings conducted on Zoom. The intervention was designed to socially integrate and support individuals in rural communities by implementing a remote, intense, three-session behavioral change protocol based on setting, implementing, and adjusting goals concerning social relationships (i.e., reducing isolation, reconnecting with and repairing lost connections, and seeking and providing help) and health (i.e., mental and physical concerns including substance use).  Following expert annotation, the recordings were systematically segmented to create our final dataset. This resulted in 517 discrete 60-second data samples, comprising 358 positive samples (requiring intervention) and 159 negative samples (not requiring intervention).

\subsubsection*{Participants}
Participants in the group meetings were individuals engaged in a program focused on community integration, health promotion, and reduction of substance-use stigma. The sessions were led by facilitators who had at least a college degree in behavioral science. The facilitators are trained in managing sensitive discussions over an 8-hour course, individual supervision, and regular support meetings. The facilitators and all research team members who had contact with the data completed CITI (Collaborative Institutional Training Initiative) certification in ethical human-subjects research. The project was approved by the University of Pennsylvania’s Institutional Review Board.

\subsubsection*{Integrated Robot System}
The architecture of our system consists of several interconnected modules that process data sequentially (Fig. 3), all designed to operate in a HIPAA-compliant environment. The process begins with the robot capturing audio from the facilitator’ side, which is streamed to a secure backend server. On the server, preprocessing is performed to video clips are transcribed using the OpenAI Whisper model to generate time-stamped text. Next, the concept modeling module applies a Concept Extractor agent, powered by a large language model (GPT-4), to parse each transcript segment. This agent populates a predefined set of binary, numeric, and ordinal concepts representing group dynamics. To maintain temporal context, a Context Integrator agent performs iterative meeting summarization, continuously updating a cumulative summary that incorporates the latest transcript and concepts. The core predictive task, determining whether an intervention is needed, is handled by a logistic regression classifier that uses the extracted concept vector as input to output a binary intervention decision. If an intervention is predicted, the action suggestion module is triggered. Here, an Intervention Advisor agent generates a context-sensitive recommendation for the facilitator, based on the cumulative summary, procedural goals, current transcript, and few-shot examples of expert interventions. The final suggestion is then sent back to the robot and delivered to the facilitator via text message and synthesized speech.

\subsubsection*{Robot Interaction Design}
The system is designed for a human-robot collaboration scenario where a physically co-located robot assists a human facilitator who is managing an online group. The robot actively listens to meeting progress and calls the server to first transcribe all audio by a locally deployed OpenAI Whisper model. The transcriptions are forwarded to later prediction and recommendation modules. Upon identifying a need for intervention, the robot presents the suggestion to the human facilitator. When an intervention is identified, the robot presents a suggestion to the human facilitator. The message is displayed on the robot’s screen and simultaneously spoken aloud, giving the facilitator the choice of receiving support via text, audio, or both. The robot’s audible output is generated using Microsoft's Edge text-to-speech (TTS) engine. This design ensures the robot acts as a ``cognitive assistant,'' supporting the facilitator without disrupting the group’s natural dynamics.

\subsubsection*{Data Collection and Feature Extraction}
The raw data consists of video recordings. These were first transcribed to text using the Whisper model. A social science expert performed qualitative coding on the transcripts to annotate the ground-truth timestamps and provide rationales for intervention. To generate training samples, we extracted 60-second video and transcript clips. Positive samples were defined as the 60-second window immediately preceding an expert-annotated intervention time. Negative samples were created by partitioning long, continuous periods ($\ge$ 5 minutes) with no intervention codes into non-overlapping 60-second chunks. The core feature extraction was performed by the Concept Extractor agent. For each 60-second transcript, the agent used GPT-4 to generate a structured feature vector composed of predefined binary, numeric, and ordinal concepts. This vector of human-interpretable concepts serves as the direct input to the downstream prediction model.

\subsection*{Statistical analysis}
We conducted statistical analyses and visualizations with Python standard scientific computing libraries. For the quantitative evaluations in Fig.~\ref{fig:quantitative} and Fig.~\ref{fig:feature_analysis}, to ensure our performance metrics were not skewed by a favorable but random split of the data, we employed a 5-fold cross-validation scheme. This method provides a more reliable estimate of the model's performance on unseen data by training and validating on different subsets of the dataset. The use of a comprehensive suite of metrics—including Accuracy, Precision, Recall, F1-score, and AUC—was deliberate, as each provides a different view of the model's behavior. The comparison against averaged results from multiple runs of the GPT baselines accounts for the inherent stochasticity of large language models, ensuring a fair and rigorous comparison.

%%%%%%%%%%%%%%%% REFERENCES %%%%%%%%%%%%%%%

\clearpage % Clear all remaining figures and tables then start a new page

% The list of references goes after the main text and before the acknowledgements
% When preparing an initial submission, we recommend you use BibTeX, like this:
%
\bibliography{science_template} % for a file named science_template.bib
\bibliographystyle{sciencemag}

\newpage

%%%%%%%%%%%%%%%% START OF SUPPLEMENT %%%%%%%%%%%%%%%

% Figures, tables, equations and pages in the supplement are numbered S1, S2 etc.
\renewcommand{\thefigure}{S\arabic{figure}}
\renewcommand{\thetable}{S\arabic{table}}
\renewcommand{\theequation}{S\arabic{equation}}
\renewcommand{\thepage}{S\arabic{page}}
\setcounter{figure}{0}
\setcounter{table}{0}
\setcounter{equation}{0}
\setcounter{page}{1} % not 0 as \newpage already started a supplementary page

\clearpage % Clear all remaining figures and tables then start a new page

% \paragraph{Caption for Movie S1.}
% \textbf{All captions must start with a short bold sentence, acting as a title.}
% Then explain what is shown in the supplementary video file.
% Give as much detail as you would for a figure e.g. explain axes, color maps etc.
% If the video is an animated equivalent of one of the static figures, state e.g.
% `Animated version of Figure~\ref{fig:example}.'

% \paragraph{Caption for Data S1.}
% \textbf{All captions must start with a short bold sentence, acting as a title.}
% Then explain what is included in the supplementary data file.
% Give as much detail as you would for a table e.g. explain the meaning of every column,
% units used, any special notation etc.

%%%%%%%%%%%%%%%% SUPPLEMENTARY REFERENCES %%%%%%%%%%%%%%%

% Do NOT include a reference list in the supplement.
% All references must be in a single list at the end of the main text.
% The copyeditors will ensure that the correct reference list appears with each version of the paper
% (print, HTML, PDF, mobile app, metadata for bibliographic databases etc.)

\end{document}